\documentclass[11pt,a4paper]{article}
\usepackage[hyperref]{acl2020}
\usepackage{times}
\usepackage{latexsym}
\usepackage{graphicx}

\usepackage{url}
\usepackage[T1]{fontenc}

\usepackage{amsmath}
\usepackage{multirow}
\usepackage{bm}
\usepackage{booktabs}

\newcommand{\tabincell}[2]{\begin{tabular}{@{}#1@{}}#2\end{tabular}}

\usepackage{microtype}

\aclfinalcopy 

\title{KdConv: A Chinese Multi-domain Dialogue Dataset Towards \\Multi-turn Knowledge-driven Conversation}

\author{Hao Zhou\thanks{\quad Equal contribution}, Chujie Zheng$^*$, Kaili Huang, Minlie Huang\thanks{\quad Corresponding author: Minlie Huang.}, Xiaoyan Zhu\\
Conversational AI Group, AI Lab., Dept. of Computer Science, Tsinghua University \\
Beijing National Research Center for Information Science and Technology, China \\
  \texttt{tuxchow@gmail.com, chujiezhengchn@gmail.com, aihuang@tsinghua.edu.cn}}

\date{}

\begin{document}
\maketitle
\begin{abstract}
The research of knowledge-driven conversational systems is largely limited due to the lack of dialog data which consist of multi-turn conversations on multiple topics and with knowledge annotations. 
In this paper, we propose a Chinese multi-domain knowledge-driven conversation dataset, \textbf{KdConv}, which grounds the topics in multi-turn conversations to knowledge graphs.
Our corpus contains 4.5K conversations from three domains (film, music, and travel), and 86K utterances with an average turn number of 19.0. These conversations contain in-depth discussions on related topics and natural transition between multiple topics.
To facilitate the following research on this corpus, we provide several benchmark models. 
Comparative results show that the models can be enhanced by introducing background knowledge, yet there is still a large space for leveraging knowledge to model multi-turn conversations for further research.
Results also show that there are obvious performance differences between different domains, indicating that it is worth to further explore transfer learning and domain adaptation. The corpus and benchmark models are publicly available\footnote{\url{https://github.com/thu-coai/KdConv}}.
\end{abstract}

\section{Introduction}
It has been a long-term goal of artificial intelligence to deliver human-like conversations, where background knowledge plays a crucial role in the success of conversational systems~\cite{Shang2015Neural,li2015diversityMMI,2017Shao-longdiverse}. In task-oriented dialog systems, background knowledge is defined as slot-value pairs, which provides key information for question answering or recommendation, and has been well defined and thoroughly studied \cite{Wen2015Semantically,zhou2016context}. In open-domain conversational systems, it is important but challenging to leverage background knowledge, which is represented as either knowledge graphs \cite{wen2017,zhou2018commonsense} or unstructured texts \cite{ghazvininejad2018knowledge}, for making effective interactions. 

Recently, a variety of knowledge-grounded conversation corpora have been proposed \cite{zhou2018dataset,dinan2018wizard,moghe2018towards,moon2019opendialkg,wu-etal-2019-proactive,liu2018knowledge,tuan2019dykgchat,qin2019conversing} to fill the gap where previous datasets do not provide knowledge grounding of the conversations \cite{godfrey1992switchboard,Shang2015Neural,lowe2015ubuntu}. 
CMU DoG \cite{zhou2018dataset}, India DoG \cite{moghe2018towards}, and Wizard of Wikipedia \cite{dinan2018wizard} demonstrate attempts for generating informative responses with topic-related Wikipedia articles. However, these datasets are not suitable for modeling topic transition or knowledge planning through multi-turn dialogs based on the relations of topics. OpenDialKG \cite{moon2019opendialkg} and DuConv \cite{wu-etal-2019-proactive} use knowledge graphs as knowledge resources. Nevertheless, the number of topics is limited to one \cite{moon2019opendialkg} or two \cite{wu-etal-2019-proactive}, which is not sufficient for diversified topic transition in human-like conversations. Therefore, these knowledge-grounded dialog datasets still have limitations in modeling knowledge interactions\footnote{Refer to knowledge planning, knowledge grounding, knowledge adaptations in dialog systems.} in multi-turn conversations.

\begin{figure*}[!t]
\centering
\includegraphics[width=16cm]{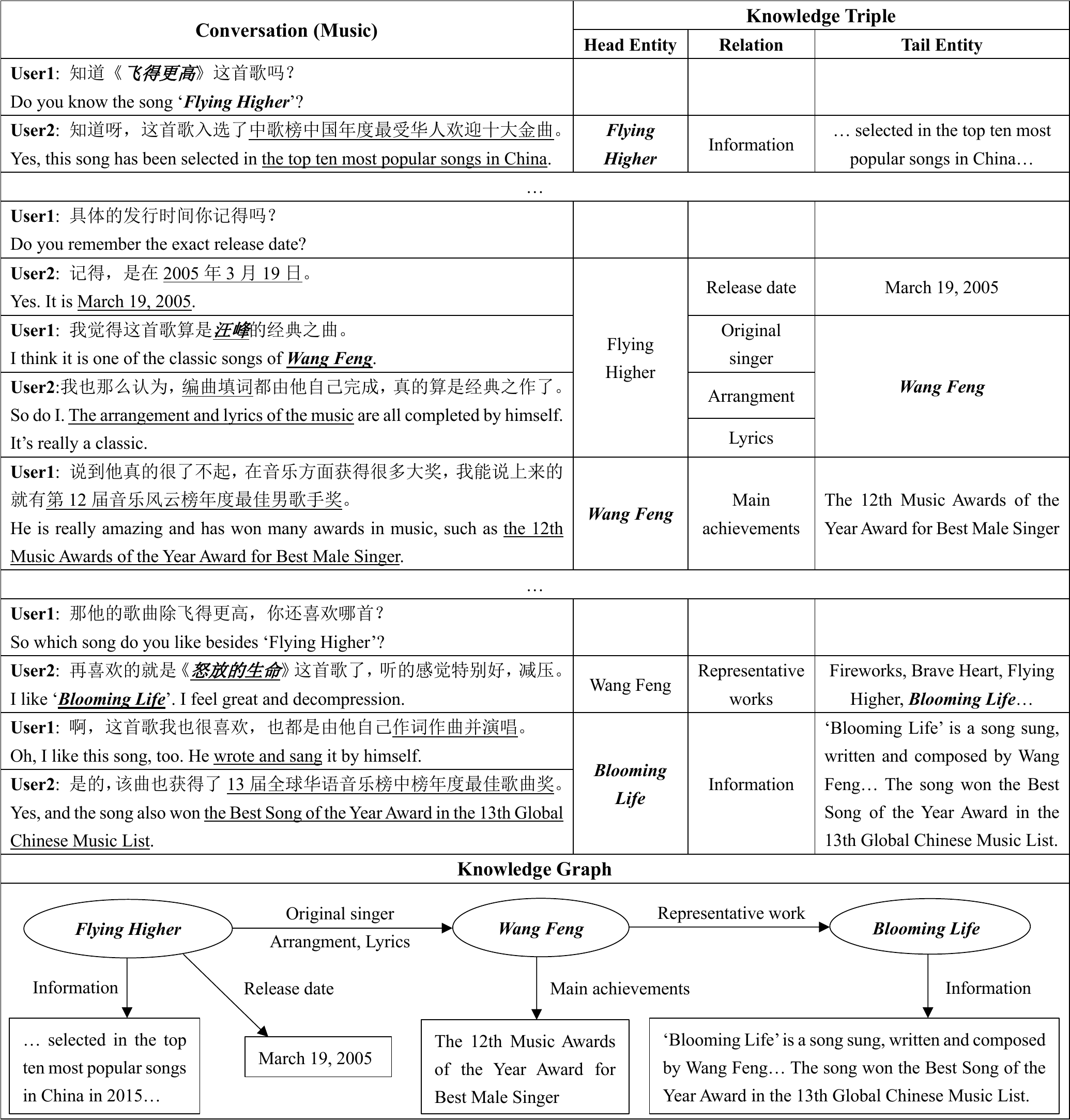}
\caption{An example in KdConv from the \textbf{music} domain. The \underline{underlined} text is the related knowledge that is utilized in conversation.
The \underline{\textit{\textbf{italic}}} text and circles are topics (refer to the distinct head entities in the knowledge triples and the central nodes with degree greater than 1 in the knowledge graph) in this dialogue. }
\label{fig:example}
\end{figure*}

In this paper, we propose \textbf{KdConv}, a Chinese multi-domain dataset towards multi-turn \textbf{K}owledge-\textbf{d}riven \textbf{Conv}ersation, which is suitable for modeling knowledge interactions in multi-turn human-like dialogues, including knowledge planning, knowledge grounding, knowledge adaptations, etc. KdConv contains 86K utterances and 4.5K dialogues in three domains, 1.5K dialogues for each domain (an example is shown in Figure \ref{fig:example}). Each utterance is annotated with related knowledge facts in the knowledge graph, which can be used as supervision for knowledge interaction modeling. 
Furthermore, conversations of KdConv contain diversified topics ranged from one to four, without any pre-defined goals or constraints, which are closer to real human-human conversations than other datasets. The relations of topics are explicitly defined in the knowledge graph. Moreover, KdConv covers three domains, including film, music, and travel, which can be used to explore knowledge adaptation between different domains.  We provide a benchmark to evaluate both generation- and retrieval-based conversational models on the proposed dataset with/without access to the corresponding knowledge. Results show that knowledge grounding contributes to the improvement of these models, while existing models are still not strong enough to deliver knowledge-coherent conversations, indicating a large space for future work.

\begin{table*}[t]
  \centering
    \scalebox{0.78}{
        \begin{tabular}{lccclccr}
        \toprule
        \textbf{Dataset} & \textbf{Language} & \textbf{Knowledge Type}  & \textbf{Annotation Level} & \textbf{Domain} & \textbf{Avg. \# turns} & \textbf{Avg. \# topics} & \textbf{\# uttrs}\\
        \midrule
        \tabincell{l}{CMU DoG} & English & Text & Sentence & Film & 22.6 & 1.0 & 130K \\ 
        \tabincell{l}{WoW}   & English & Text &  Sentence & Multiple & 9.0 & 2.0 & 202K \\
        \tabincell{l}{India DoG} & English  & Text \& Table & Sentence & Film & 10.0 & 1.0 & 91K \\
        \tabincell{l}{OpenDialKG
        } & English & Graph & Sentence & \tabincell{l}{Film, Book,\\Sport, Music} & 5.8 & 1.0 & 91K \\
        \midrule
        \tabincell{l}{DuConv} & Chinese & Text \& Graph & Dialog & Film  & 9.1 & 2.0 & 270K \\
        \cmidrule{1-8}
        \textbf{KdConv (ours)} & \textbf{Chinese} & \textbf{Text \& Graph} & \textbf{Sentence} & \tabincell{l}{\textbf{Film, Music,}\\\textbf{Travel}} & \textbf{19.0} & \textbf{2.3} & \textbf{86K} \\
        \bottomrule
        \end{tabular}
    }
  \caption{Comparison between our corpus and other human-labeled knowledge-grounded dialogue corpora.}
  \label{tab:comparison}%
\end{table*}

In summary, this paper makes the following contributions:
\begin{itemize}
\item We collect a new dataset, KdConv, for knowledge-driven conversation generation in Chinese. KdConv contains 86K utterances and 4.5K dialogues in three domains (film, music, and travel). The average turn number is about 19, remarkably longer than those in other corpora. 

\item KdConv provides a benchmark to evaluate the ability of generating conversations with access to the corresponding knowledge in three domains. The corpus can empower the research of not only knowledge-grounded conversation generation, but also domain adaptation or transfer learning between similar domains (e.g., from film to music) or dissimilar domains (e.g., from music to travel).

\item We provide benchmark models on this corpus to facilitate further research, and conduct extensive experiments. Results show that the models can be enhanced by introducing background knowledge, but there is still much room for further research. The corpus and the models are publicly available\footnote{\url{https://github.com/thu-coai/KdConv}}.

\end{itemize}

\section{Related Work} 
\label{sec:related_work}

Recently, open-domain conversation generation has been largely advanced due to the increase of publicly available dialogue data \cite{godfrey1992switchboard,ritter2010unsupervised,Shang2015Neural,lowe2015ubuntu}. However, the lack of annotation of background information or related knowledge results in significantly degenerated conversations, where the text is bland and strangely repetitive \cite{holtzman2019curious}.
These models produce conversations that are substantially different from those humans make, which largely rely on background knowledge. 

To facilitate the development of conversational models that mimic human conversations, there have been several knowledge-grounded corpora proposed. 
Some datasets \cite{zhou2018dataset,ghazvininejad2018knowledge,liu2018knowledge,tuan2019dykgchat,qin2019conversing} collect dialogues and label the knowledge annotations using NER, string match, artificial scoring, and filtering rules based on external knowledge resources \cite{liu2018knowledge}. However, mismatches between dialogues and knowledge resources introduce noises to these datasets. 
To obtain the high-quality knowledge-grounded datasets, some studies construct dialogues from scratch with human annotators, based on the unstructured text or structured knowledge graphs.
For instance, several datasets \cite{zhou2018dataset,dinan2018wizard,Gopalakrishnan2019} have human conversations where one or both participants have access to the unstructured text of related background knowledge, while OpenDialKG \cite{moon2019opendialkg} and DuConv \cite{wu-etal-2019-proactive} build up their corpora based on structured knowledge graphs. In Table \ref{tab:comparison}, we present a survey on existing human-labeled knowledge-grounded dialogue datasets.

CMU DoG \cite{zhou2018dataset} utilizes 30 Wikipedia articles about popular movies as grounded documents, which explores two scenarios: only one participant has access to the document, or both have. 
Also using Wikipedia articles, however, Wizard of Wikipedia (WoW) \cite{dinan2018wizard} covers much more dialogue topics (up to 1,365), which puts forward a high demand for the generalization ability of dialog generation models.
One other difference from CMU DoG is that in WoW, only one participant has access to an information retrieval system that shows the worker paragraphs from Wikipedia possibly relevant to the conversation, which is unobservable to the other.
In addition to the unstructured text, India DoG \cite{moghe2018towards} uses fact tables as background resources.

The idea of using structured knowledge to construct dialogue data is also adopted in OpenDialKG \cite{moon2019opendialkg}, which has a similar setting to KdConv. 
OpenDialKG contains chit-chat conversations between two agents engaging in a dialog about a given topic. 
It uses the Freebase knowledge base \cite{bast2014easy} as background knowledge. In OpenDialKG, the entities and relations that are mentioned in the dialog are annotated, and it also covers multiple domains (film, books, sports, and music). However, the limitation is that there are much fewer turns in a conversation, and the whole dialogue is restricted to only one given topic, which is not suitable for modeling topic transition in human-like conversations. 

To the best of our knowledge, DuConv \cite{wu-etal-2019-proactive} is the only existing Chinese human-labeled knowledge-grounded dialogue dataset. 
DuConv also utilizes unstructured text like short comments and synopsis, and structured knowledge graphs as knowledge resources. 
Given the knowledge graph, it samples two linked entities, one as the transitional topic and the other as the goal topic, to construct a conversation path. 
This path is used to guide participants toward the goal of the dialogue, which, as argued in \citet{wu-etal-2019-proactive}, can guide a model to deliver proactive conversations. 
However, the existence of the target path is inconsistent with an open dialogue in reality because humans usually do not make any assumption about the final topic of a conversation. 
Beyond that, the knowledge graph and the goal knowledge path are only annotated for the whole dialogue, which cannot provide explicit supervision on knowledge interactions for conversational models.

\begin{table*}[t]
  \centering
  \scalebox{1.0}{
    \begin{tabular}{l|rrr|r}
    \toprule
    \textbf{Domain} & \multicolumn{1}{c}{\textbf{Film}} & \multicolumn{1}{c}{\textbf{Music}} & \multicolumn{1}{c|}{\textbf{Travel}} & \multicolumn{1}{c}{\textbf{Total}} \\
    \midrule
    \textbf{\# entities } & 7,477 & 4,441 & 1,154 & 13,072 \\
    \ \ \textbf{ (\# start/\# extended)} & \multicolumn{1}{l}{(559/6,917)} & \multicolumn{1}{l}{(421/4,020)} & \multicolumn{1}{l|}{(476/678)} & \multicolumn{1}{l}{(1,456/11,615)} \\
    \textbf{\# relations} & 4,939 & 4,169 & 7     & 9,115 \\
    \textbf{\# triples} & 89,618 & 56,438 & 10,973 & 157,029 \\
    \midrule
    \textbf{Avg. \# triples per entity} & 12.0  & 12.7  & 9.5   & 12.0  \\
    \textbf{Avg. \# tokens per triple} & 20.5  & 19.2  & 20.9  & 20.1  \\
    \textbf{Avg. \# characters per triple} & 51.6  & 45.2  & 39.9  & 48.5  \\
    \bottomrule
    \end{tabular}%
}
  \caption{Statistics of the knowledge graphs used in constructing KdConv.}
  \label{tab:graph}%
\end{table*}%

\begin{table*}[t]
  \centering
  \scalebox{1.0}{
    \begin{tabular}{l|rrr|r}
    \toprule
    \textbf{Domain} & \multicolumn{1}{c}{\textbf{Film}} & \multicolumn{1}{c}{\textbf{Music}} & \multicolumn{1}{c|}{\textbf{Travel}} & \multicolumn{1}{c}{\textbf{Total}}\\
    \midrule
    \textbf{\# dialogues} & \multicolumn{3}{c|}{1,500} & 4,500\\
    \textbf{\# dialogues in Train/Dev/Test} & \multicolumn{3}{c|}{1,200/150/150} & 3,600/450/450 \\
    \textbf{\# utterances} & 36,618 & 24,885 & 24,093 & 85,596 \\
    \textbf{Avg. \# utterances per dialogue} & 24.4  & 16.6  & 16.1  & 19.0 \\
    \textbf{Avg. \# topics per dialogue} & 2.6 & 2.1 & 2.2 & 2.3 \\
    \cmidrule{1-5}
    \textbf{Avg. \# tokens per utterance} & 13.3  & 12.9  & 14.5  & 13.5 \\
    \textbf{Avg. \# characters per utterance} & 20.4  & 19.5  & 22.9  & 20.8 \\
    \textbf{Avg. \# tokens per dialogue} & 323.9 & 214.7 & 233.5 & 257.4 \\
    \textbf{Avg. \# characters per dialogue} & 497.5 & 324.0 & 367.8 & 396.4 \\
    \cmidrule{1-5}
    \textbf{\# entities} & 1,837 & 1,307 & 699   & 3,843 \\
    \textbf{\# start entities} & 559   & 421   & 476   & 1,456 \\
    \textbf{\# relations} & 318   & 331   & 7     & 656 \\
    \textbf{\# triples} & 11,875 & 5,747 & 5,287 & 22,909 \\
    \cmidrule{1-5}
    \textbf{Avg. \# triples per dialogue} & 16.8  & 10.4  & 10.0  & 10.1 \\
    \textbf{Avg. \# tokens per triple} & 25.8  & 29.7  & 31.0  & 28.3 \\
    \textbf{Avg. \# characters per triple} & 49.4  & 56.8  & 57.4  & 53.6 \\
    \bottomrule
    \end{tabular}%
}
  \caption{Statistics of KdConv.}
  \label{tab:statistics}%
\end{table*}%

\section{Dataset Collection}

KdConv is designed to collect open-domain multi-turn conversations for modeling knowledge interactions in human-like dialogues, including knowledge planning, knowledge grounding, knowledge adaptations, etc.
However, the open-domain background or commonsense knowledge is too large in scale (e.g., there are over 8 million concepts and 21 million relations in ConceptNet \cite{speer2013conceptnet}). Thus, it is costly and time-consuming to collect multi-turn conversations from scratch based on such large-scale knowledge. KdConv is proposed as one small step to achieve this goal, where we narrowed down the scale of background knowledge to several domains (film, music, and travel) and collected conversations based on the domain-specific knowledge. KdConv contains similar domains (film and music) and dissimilar domains (film and travel) so that it offers the possibility to investigate the generalization and transferability of knowledge-driven conversational models with transfer learning or meta learning\cite{gu2018meta,mi2019meta}.

In the following subsections, we will describe the two steps in data collection: (1) Constructing the domain-specific knowledge graph; (2) Collecting conversation utterances and knowledge interactions by crowdsourcing.

\subsection{Knowledge Graph Construction}

As the sparsity and the large scale of the knowledge were difficult to handle, we reduced the range of the domain-specific knowledge by crawling the most popular films and film stars, music and singers, and attractions as start entities, from several related websites for the film\footnote{\url{https://movie.douban.com/top250}}/music\footnote{\url{https://music.douban.com/top250}}/travel\footnote{\url{https://travel.qunar.com/p-cs299914-beijing-jingdian}} domain. The knowledge of these start entities contains both structured knowledge triples and unstructured knowledge texts, which make the task more general but challenging. After filtering the start entities which have few knowledge triples, 
the film/music/travel domain contains 559/421/476 start entities, respectively.

After crawling and filtering the start entities, we built the knowledge graph for each domain. Given the start entities as seed, we retrieved their neighbor entities within three hops from XLORE, a large-scale English-Chinese bilingual knowledge graph \cite{wang2013xlore}. We merged the start entities and these retrieved entities (nodes in the graph) and relations (edges in the graph) into a domain-specific knowledge graph for film and music domains. For the travel domain, we built the knowledge graph with the knowledge crawled only from the Web, because XLORE provides little knowledge for start entities in the travel domain. There are two types of entities in the knowledge graph: one is the start entities crawled from the websites, the other is the extended entities that are retrieved from XLORE (film/music), or websites (travel) to provide related background knowledge. The statistics of the knowledge graphs used in constructing KdConv are provided in Table \ref{tab:graph}.

\subsection{Dialogue Collection}
We recruited crowdsourced annotators to generate multi-turn conversations that are related to the domain-specific knowledge graph without any pre-defined goals or constraints. 
During the conversation, two speakers both had access to the knowledge graph rather than that only one participant had access to the knowledge, as proposed in WoW \cite{dinan2018wizard} where one party always leads the conversation with an expert-apprentice mode. Allowing two participants to access the knowledge, in our corpus the two parties can dynamically change their roles, as either leader or follower, which is more natural and real to human conversations.
In addition to making dialogue utterances, the annotators were also required to record the related knowledge triples if they generated an utterance according to some triples. To increase the knowledge exposure in the collected conversations, the annotators were instructed to start the conversation based on one of the start entities, and they were also encouraged to shift the topic of the conversation to other entities in the knowledge graph. Thus, the topics of conversations and the knowledge interactions in KdConv are diversified and unconstrained. In order to ensure the naturalness of the generated conversations, we filtered out low-quality dialogues, which contain grammatical errors, inconsistencies of knowledge facts, etc. The distinct-4 score is 0.54/0.51/0.42 for the film/music/travel domain, which is comparable to the score of DuConv \cite{wu-etal-2019-proactive}, 0.46. The distinct-4 score decreases, due to the decrease of knowledge triples and utterances in three domains, as shown in Table \ref{tab:statistics}.

\begin{figure}[!t]
    \centering
    \includegraphics[width=0.8\linewidth]{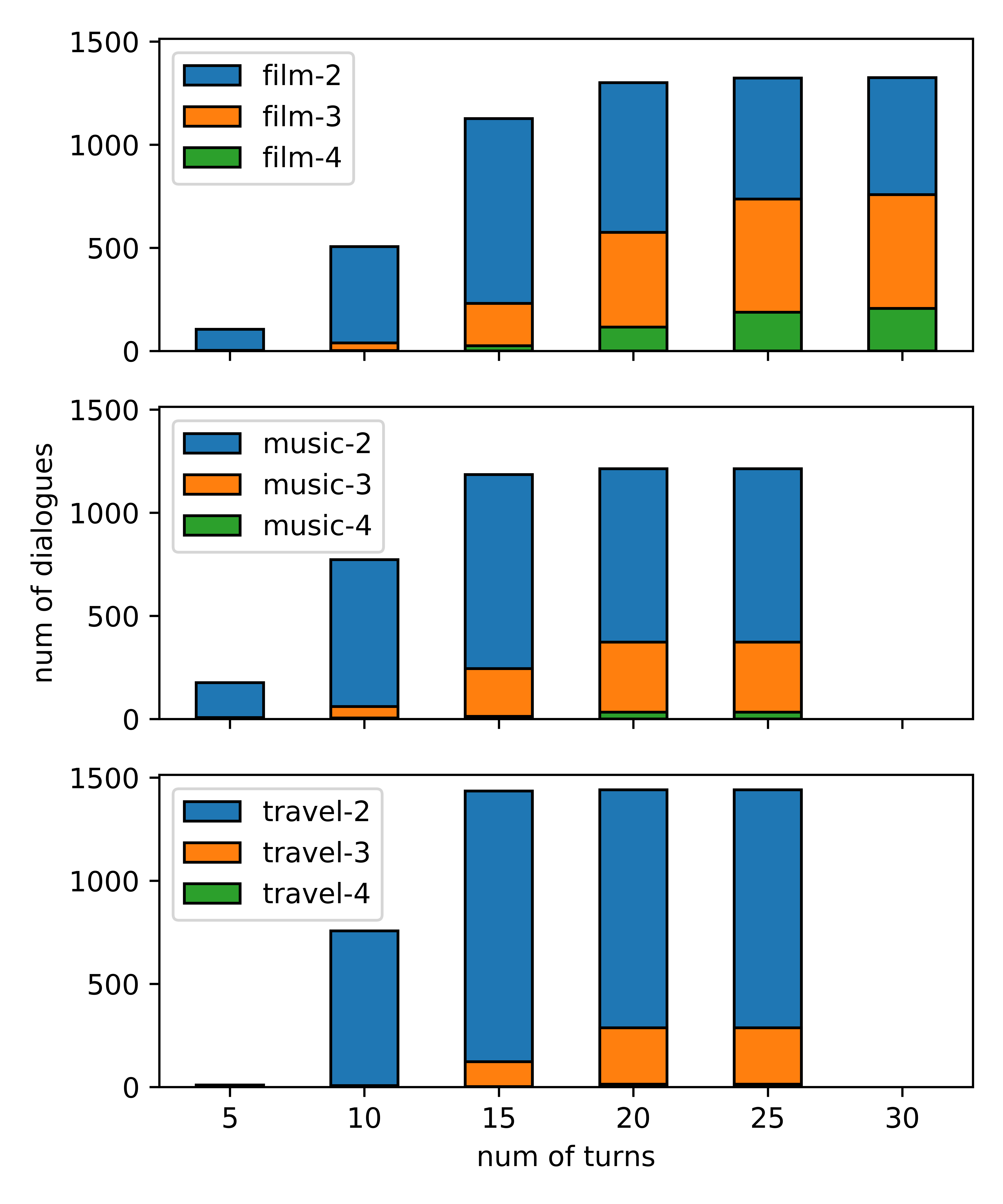}
    \caption{Statistics of the number of dialogues where at least $k (k=2,3,4)$ topics have been discussed in the first $n$ turns. The proportions of dialogues that contain 3 or 4 topics become larger when the dialog turn becomes longer. }
    \label{fig:turn2topic}
\end{figure}

\subsection{Corpus Statistics}

The detailed statistics of KdConv are shown in Table \ref{tab:statistics}. We collect 1,500 dialogues for each domain. The training, validation, and test sets are partitioned with the ratio of 8:1:1. Note that the number of conversation turns in the film domain is larger than those in the music/travel domains (24.4 vs. 16.6/16.1), while the utterance lengths are similar (13.3 vs. 12.9/14.5 at the token level, and 20.4 vs. 19.5/22.9 at character level). As aforementioned, the dialogues in the real world are not limited to one or two topics, while discussing multiple topics in depth usually requires a conversation having enough number of turns. In order to verify this point, we analyze the relationship between the number of turns and the number of topics. Note that the topics are defined as the distinct head entities in the knowledge triples and the central nodes with a degree greater than 1 in the knowledge graph.

The results of three domains are shown in Figure \ref{fig:turn2topic}. 
Given a number $k (k=2,3,4)$ of topics and a number $n$ of conversation turns, we count the number of dialogues where at least $k$ topics have been discussed in the first $n$ turns. 
It can be observed that more topics tend to appear in a dialogue only if there are enough conversation turns. 
For instance, most dialogues involve at least 2 topics when the number of turns exceeds 15. 
This is consistent with the fact that if a conversation is very short, speakers will not be able to discuss in detail, let alone natural transition between multiple topics.

\begin{table}[htbp]
  \centering
  \scalebox{0.75}{
    \begin{tabular}{c|l}
    \toprule
    \multicolumn{2}{c}{\textbf{Topic Transition}} \\
    \midrule
    \multirow{3}[2]{*}{1 Hop} & $T_1 -$Major Work$\rightarrow T_2$ \\
          & $T_1 -$Star$\rightarrow T_2$ \\
          & $T_1 -$Director$\rightarrow T_2$ \\
    \midrule
    \multirow{3}[2]{*}{2 Hop} & $T_1 -$Major Work$\rightarrow T_2 -$Star$\rightarrow T_3$ \\
          & $T_1 -$Major Work$\rightarrow T_2 -$Director$\rightarrow T_3$ \\
          & $T_1 -$Star$\rightarrow T_2 -$Major Work$\rightarrow T_3$ \\
    \midrule
    \multirow{3}[2]{*}{3 Hop} & $T_1 -$Major Work$\rightarrow T_2 -$Star$\rightarrow T_3 -$Major Work$\rightarrow T_4$ \\
          & $T_1 -$Star$\rightarrow T_2 -$Major Work$\rightarrow T_3 -$Director$\rightarrow T_4$ \\
          & $T_1 -$Major Work$\rightarrow T_2 -$Star$\rightarrow T_3 -$Information$\rightarrow T_4$ \\
    \bottomrule
    \end{tabular}%
  }
  \caption{Top-3 topic transition of the film domain, where $T_n$ denotes the $n$-th topic of a dialog and $T_n-X\rightarrow T_{n+1}$ represents the relation $X$ between $T_n$ and $T_{n+1}$.}
  \label{tab:transition}%
\end{table}%

To analyze topic transition in our dataset, we provide top-3 topic transition in the film domain, as shown in Table \ref{tab:transition}. As can be seen, topic transition has diverse patterns conditioned on different hops. With the increase of the hops of topic transition, the complexity of topic transition goes up. Compared to DuConv \cite{wu-etal-2019-proactive}, the dialogues of KdConv contain multiple and diverse topics instead of fixed two topics, leading to diverse and complex topic transition, which are more suitable for the research of knowledge planning in human-like conversations. Note that the relation ``$-$Information$\rightarrow$'' appeared in the last row is different from the other relations, which means the target topic is mentioned in unstructured texts describing the information about the source topic. The low frequency of the relation ``$-$Information$\rightarrow$'' demonstrates that people prefer to shift the topic according to the structured relations rather than unstructured texts, as adopted in WoW \cite{dinan2018wizard}. 

\section{Experiments}

\subsection{Models}
To provide benchmark models for knowledge-driven conversation modeling, we evaluated both generation- and retrieval-based models on our corpus. In order to explore the role of knowledge annotation, we evaluated the models with/without access to the knowledge graph of our dataset.

\subsubsection{Generation-based Models}
\noindent \textbf{Language Model (LM)} \cite{bengio2003neural}: We trained a language model that maximizes the log likelihood: $\log \mathcal{P}(\bm{x}) = \sum_{t}\log\mathcal{P}(x_t|x_{<t})$,
where $\bm{x}$ denotes a long sentence that sequentially concatenates all the utterances of a dialogue.

\noindent \textbf{Seq2Seq} \cite{sutskever2014sequence}: An encoder-decoder model augmented with attention mechanism \cite{bahdanau2014neural}. 
The input of the encoder was the concatenation of the past $k-1$ utterances, while the target output of the decoder was the $k$-th utterance. $k$ was set to 8 in the experiment. If there were fewer than $k-1$ sentences in the dialogue history, all the past utterances would be used as input.

\noindent \textbf{HRED} \cite{Serban2016Building}: A hierarchical recurrent encoder-decoder model that has a specific context RNN to incorporate historical conversational utterances into a context state, which is used as the initial hidden state of the decoder.
The adapted model generates the $k$-th utterance based on the past $k-1$ utterances, where $k$ was also set to 8, for fair comparison with Seq2Seq. 

All the generative models were trained by optimizing the cross-entropy loss:
\begin{align*}
    \mathcal{L}^{(g)}_0 = -\frac{1}{T}\sum_{t=1}^T\log \mathcal{P}(\hat{x}_t=x_t),
\end{align*}
where $\hat{x}_t$ denotes the predicted token at the time step $t$, while $x_t$ is the $t$-th token of the target sentence.

\subsubsection{Retrieval-based Model} 
\label{sec:retrieve-bert}

\noindent \textbf{BERT} \cite{devlin-etal-2019-bert}: We adapted this deep bidirectional transformers \cite{vaswani2017attention} as a retrieval-based model. 
For each utterance (except the first one in a dialog), we extracted keywords in the same way as \citet{wu-etal-2017-sequential} and retrieved 10 response candidates, including the golden truth based on the BM25 algorithm \cite{robertson1995okapi}.
The training task is to predict whether a candidate is the correct next utterance given the context, where a sigmoid function was used to output the probability score $\hat{y}=\mathcal{P}(y=1)$ and the cross-entropy loss was optimized: 
\begin{align*}
    \mathcal{L}^{(r)}_0 = -y\log \hat{y}-(1-y)\log (1-\hat{y}),
\end{align*}
where $y\in\{0,1\}$ is the true label.
For the test, we selected the candidate response with the largest probability.

\subsubsection{Knowledge-aware Models}
A key-value memory module \cite{miller2016key} is introduced to the aforementioned models 
to utilize the knowledge information. 
We treated all knowledge triples mentioned in a dialogue as the knowledge information in the memory module. For a triple that is indexed by $i$, we represented the key memory and the value memory respectively as a key vector $\bm{k}_i$ and a value vector $\bm{v}_i$, where $\bm{k}_i$ is the average word embeddings of the head entity and the relation, and $\bm{v}_i$ is those of the tail entity. We used a query vector $\bm{q}$ to attend to the key vectors $\bm{k}_i (i=1,2,...)$: $\alpha_i= \mathrm{softmax}_i(\bm{q}^T\bm{k}_i)$, then the weighted sum of the value vectors $\bm{v}_i (i=1,2,...)$, $\bm{v}=\sum_{i}\alpha_i \bm{v}_i$, was incorporated into the decoding process (for the generation-based models, concatenated with the initial state of the decoder) or the classification (for the retrieval-based model, concatenated with the {\tt <CLS>} vector).
For Seq2Seq, $\bm{q}$ is the final hidden state of the encoder. For HRED, we treated the context vector as the query, while for BERT, the output vector of {\tt <CLS>} was used.

Note that our dataset has a sentence-level annotation on the knowledge triples that each utterance uses. To force the knowledge-aware models to attend to the golden KG triples, we added an extra attention loss (for retrieval-based models, this loss was computed only on the positive examples):
\begin{align*}
    \mathcal{L}_\mathrm{att} = -\frac{1}{|\{\mathrm{truth}\}|}\sum_{i\in \{\mathrm{truth}\}} \log\alpha_i,
\end{align*}
where $\{\mathrm{truth}\}$ is the set of indexes of triples that are used in the true response. The total loss are the weighted sum of $\mathcal{L}^{(l)}_0$ and $\mathcal{L}_\mathrm{att}$:
\begin{align*}
    \mathcal{L}^{(l)}_\mathrm{tot} = \mathcal{L}^{(l)}_0 + \lambda \mathcal{L}_\mathrm{att},\ \ l \in \{g, r\}.
\end{align*}

Note that the knowledge-enhanced BERT was initialized from the fine-tuned BERT discussed in Section \ref{sec:retrieve-bert}, and the parameters of the transformers were frozen during training the knowledge related modules. The purpose was to exclude the impact of the deep transformers but only examine the potential effects introduced by the background knowledge.

\begin{table*}[t]
  \centering
  \scalebox{0.8}{
    \begin{tabular}{l|cc|c|cccc|cccc}
    \toprule
    \textbf{Model} & \multicolumn{2}{c|}{\textbf{Hits@1/3}} & \multicolumn{1}{c|}{\textbf{PPL}} & \multicolumn{4}{c|}{\textbf{BLEU-1/2/3/4}} & \multicolumn{4}{c}{\textbf{Distinct-1/2/3/4}} \\
    \midrule
    \multicolumn{12}{c}{\textbf{Film}}\\
    \midrule
    \textbf{LM} & 14.30 & 35.70 & \textbf{21.91} & 24.22 & 12.40 & 7.71 & 4.27 & 2.32 & 6.13 & 10.88 & 16.14 \\
    \textbf{Seq2Seq} & 17.54 & 40.57 & 23.88 & 26.97 & 14.31 & 8.53 & 5.30 & 2.51 & 7.14 & 13.62 & 21.02 \\
    \textbf{HRED} & 16.45 & 40.62  & 24.74 & 27.03 & 14.07 & 8.30 & 5.07 & 2.55 & 7.35 & 14.12 & 21.86    \\
    \textbf{BERT} & 65.36 &	\underline{91.79} &	-&	81.64 	&77.68 	&75.47 	&73.99 	&8.55 	&31.28 	&51.29 	&63.38 \\
    \cmidrule{1-12}
    \textbf{Seq2Seq + know} & \textbf{17.77} & \textbf{41.66} & 25.56 & 27.45 & 14.51 & 8.66 & 5.32 & 2.85 & 7.98 & 15.09 & 23.17 \\
    \textbf{HRED + know} & 17.38 & 39.79 & 26.27 & \textbf{27.94} & \textbf{14.69} & \textbf{8.73} & \textbf{5.40} & \textbf{2.86} & \textbf{8.08} & \textbf{15.81} & \textbf{24.93}    \\
    \textbf{BERT + know} & \underline{65.67} &	\underline{91.79} &-	&	\underline{81.98} &	\underline{78.08} &	\underline{75.90} &	\underline{74.44} &	\underline{8.59} &	\underline{31.47} &	\underline{51.63} &	\underline{63.78}  \\
    \midrule
    \multicolumn{12}{c}{\textbf{Music}} \\
    \midrule
    \textbf{LM} & 18.09 & 39.36 & \textbf{14.61} & 25.80 & 13.93 & 8.61 & 5.57 & 2.72 & 7.31 & 12.69 & 18.64  \\
    \textbf{Seq2Seq} & 22.65 & 44.43 & 16.17 & 28.89 & 16.56 & 10.63 & 7.13 & 2.52 & 7.02 & 12.69 & 18.78  \\
    \textbf{HRED} & 21.20 & 42.84 & 16.82 & \textbf{29.92} & 17.31 & 11.17 & 7.52 & 2.71 & 7.71 & 14.07 & 20.97     \\
    \textbf{BERT} &55.64 &	\underline{86.90} &	-&	78.71 &	73.61 &	70.55 &	68.43 &	6.57 &	26.75 &	44.75 &	55.85   \\
    \cmidrule{1-12}
    \textbf{Seq2Seq + know} & \textbf{22.90} & \textbf{47.14} & 17.12 & 29.60 & 17.26 & 11.36 & 7.84 & \textbf{3.93} & \textbf{12.35} & \textbf{23.01} & \textbf{34.23}  \\
    \textbf{HRED + know} & 21.82 & 45.33 & 17.69 & 29.73 & \textbf{17.51} & \textbf{11.59} & \textbf{8.04} & 3.80 & 11.70 & 22.00 & 33.37     \\
    \textbf{BERT + know} & \underline{56.08} &	86.87 &	-&	\underline{78.98} &	\underline{73.91} &	\underline{70.87} &	\underline{68.76} &	\underline{6.59} &	\underline{26.81} &	\underline{44.84} &	\underline{55.96}  \\
    \midrule
    \multicolumn{12}{c}{\textbf{Travel}} \\
    \midrule
    \textbf{LM} & 22.16 & 41.27 & \textbf{8.86} & 27.51 & 17.79 & 12.85 & 9.86 & 3.18 & 8.49 & 13.99 & 19.91  \\
    \textbf{Seq2Seq} & 27.07 & 46.34 & 10.44 & 29.61 & 20.04 & 14.91 & 11.74 & 3.75 & 11.15 & 19.01 & 27.16  \\
    \textbf{HRED} & 25.76 & 46.11 & 10.90 & 30.92 & 20.97 & 15.61 & 12.30 & 4.15 & 12.01 & 20.52 & 28.74     \\
    \textbf{BERT} & 45.25 &	71.87 &	-&	81.12 &	76.97 &	74.47 &	72.73 &	7.17 &	22.55 &	34.03 &	40.78  \\
    \cmidrule{1-12}
    \textbf{Seq2Seq + know} & \textbf{29.67} & \textbf{50.24} & 10.62 & \textbf{37.04} & \textbf{27.28} & \textbf{22.16} & \textbf{18.94} & \textbf{4.25} & \textbf{13.64} & \textbf{24.18} & 34.08  \\
    \textbf{HRED + know} & 28.84 & 49.27 & 11.15 & 36.87 & 26.68 & 21.31 & 17.96 & 3.98 & 13.31 & 24.06 & \textbf{34.35}     \\
    \textbf{BERT + know} &\underline{45.74} &	\underline{71.91} &	-&	\underline{81.28} &	\underline{77.17} &	\underline{74.69} &	\underline{72.97} &	\underline{7.20} &	\underline{22.62} &	\underline{34.11} &	\underline{40.86}  \\
    \bottomrule
    \end{tabular}%
  }
  \caption{Automatic evaluation. The best results of generative models and retrieval models are in \textbf{bold} and \underline{underlined} respectively. ``+ know'' means the models enhanced by the knowledge base.}
  \label{tab:automatic}%
\end{table*}%

\subsection{Implementation Details}

We implemented the above models with TensorFlow\footnote{\url{https://github.com/tensorflow/tensorflow}} \cite{abadi2016tensorflow} and PyTorch\footnote{\url{https://github.com/pytorch/pytorch}} \cite{paszke2017automatic}.
The Jieba Chinese word segmenter\footnote{\url{https://github.com/fxsjy/jieba}} was employed for tokenization. 
The 200-dimensional word embeddings were initialized by \citet{song2018directional}, while the unmatched ones were randomly sampled from a standard normal distribution $\mathcal{N}(0, 1)$.
The type of RNN network units was all GRU \cite{cho-etal-2014-learning} and the number of hidden units of GRU cells were all set to 200. 
ADAM \cite{kingma2014adam} was used to optimize all the models with the initial learning rate of $5\times 10^{-5}$ for BERT and $10^{-3}$ for others.
The mini-batch sizes are set to 2 dialogues for LM and 32 pairs of post and response for Seq2Seq and HRED.

\subsection{Automatic Evaluation}

\subsubsection{Metrics}
We measured the performance of all the retrieval-based models using Hits@1 and Hits@3, same as \citet{zhang-etal-2018-personalizing} and \citet{wu-etal-2019-proactive}. \footnote{For generative models, the rank is decided by the PPL values of candidate responses.}
We adopted several widely-used metrics to measure the quality of the generated response. We calculated Perplexity (PPL) to evaluate whether the generation result is grammatical and fluent. BLEU-1/2/3/4 \cite{papineni2002bleu} is a popular metric to compute the $k$-gram overlap between a generated sentence and a reference \cite{sordoni-etal-2015-neural,li-etal-2016-diversity}. Distinct-1/2/3/4 \cite{li-etal-2016-diversity} is also provided to evaluates the diversity of generated responses.

\subsubsection{Results}
The results are shown in Table \ref{tab:automatic}. We analyze the results from the following perspectives:

\textbf{The influence of knowledge:}  after introducing the knowledge, all the models were improved in terms of all the metrics except PPL in all the domains. First, all the models obtain higher Hits@1 scores (in the music domain, BERT obtains an improvement of 0.4 on Hits@1). After incorporating the knowledge into BERT, the performance of Hits@1 improves slightly, because the memory network which models knowledge information is rather shallow, compared to the deep structure in BERT.
Second, Seq2Seq and HRED both have better BLEU-$k$ scores (in the travel domain, Seq2Seq obtains an improvement of 7.2 on BLEU-4), which means a better quality of generated responses. Third, the two generation-based models also gain larger Distinct-$k$ values (in the music domain, HRED obtains an improvement of 12.4 on Distinct-4), which indicates a better diversity of the generated results.

\textbf{Comparison between models:}
In all the three domains, the knowledge-aware BERT model achieves the best performance in most of the metrics, as it retrieves the golden-truth response at a fairly high rate. 
HRED performs best in BLEU-$k$ and Distinct-$k$ among all the generation-based baselines without considering the knowledge. 
Knowledge-aware HRED has better results of BLEU-$k$ in the film and music domains and better results of Distinct-$k$ in the film domain, while the knowledge-enhanced Seq2Seq achieves the best Hits@1/3 scores among all the generation-based models.

\textbf{Comparison between domains:} 
For retrieval-based models, the performance is best in the film domain but worst in the travel domain, largely affected by the data size (see Table \ref{tab:statistics}). For generation-based models, however, the performance improves from the film domain to the travel domain, as the average number of utterances per dialogue decreases from 24.4 in the film domain to 16.1 in the travel domain (see Table \ref{tab:statistics}). The more utterances a dialogue contains, the more difficulties in conversation modeling for generation-based models. Besides, the more diverse knowledge (1,837 entities and 318 relations in the film domain, vs. 699 entities and 7 relations in the travel domain) also requires the models to leverage knowledge more flexibly. The difference between different domains can be further explored in the setting of transfer learning or meta learning in the following research.

\begin{table}[t]
  \centering
  \scalebox{1.0}{
    \begin{tabular}{l|c|c}
    \toprule
    \textbf{Model} & \textbf{Fluency} & \textbf{Coherence} \\
    \midrule
    \textbf{Film}$\bm{\ \setminus \ \kappa}$ & 0.50 & 0.61 \\
    \cmidrule{1-3}
    \textbf{HRED} & 1.64 & 1.19   \\
    \textbf{HRED + know} & \underline{\textit{1.78}} & \underline{\textit{1.28}}  \\
    \textbf{BERT + know} & \textbf{2.00} &  \textbf{1.79}  \\
    \midrule
    \textbf{Music}$\bm{\ \setminus \ \kappa}$ & 0.37 & 0.57 \\
    \cmidrule{1-3}
    \textbf{HRED} & \underline{{1.90}}  & 1.30  \\
    \textbf{HRED + know} & 1.86  &  \underline{{1.36}}    \\
    \textbf{BERT + know} & \textbf{2.00}  & \textbf{1.80}  \\
    \midrule
    \textbf{Travel}$\bm{\ \setminus \ \kappa}$ & 0.55 & 0.74 \\
    \cmidrule{1-3}
    \textbf{HRED} & 1.77   & 1.10   \\
    \textbf{HRED + know} & 1.78 &  \underline{\textit{1.31}} \\
    \textbf{BERT + know} & \textbf{2.00}  & \textbf{1.76}  \\
    \bottomrule
    \end{tabular}%
  }
  \caption{Manual evaluation. The best results ($t$-test, $p$-value < 0.005) are in \textbf{bold}. Between two generative models, the significantly better results are \underline{\textit{italic underlined}} ($t$-test, $p$-value < 0.005) or \underline{underlined} ($t$-test, $p$-value < 0.05). $\bm{\kappa}$ is the Fleiss' kappa value. ``+ know'' means the models enhanced by knowledge information.
  }
  \label{tab:manual}%
\end{table}%

\subsection{Manual Evaluation}

To better understand the quality of the generated responses from the semantic and knowledge perspective, we conducted the manual evaluation for knowledge-aware BERT, knowledge-aware HRED, and HRED, which have achieved advantageous performance in automatic evaluation\footnote{We omitted the BERT model because it performs similarly to knowledge-aware BERT as shown in automatic evaluation.}.

\begin{figure*}[!t]
\centering
\includegraphics[width=16cm]{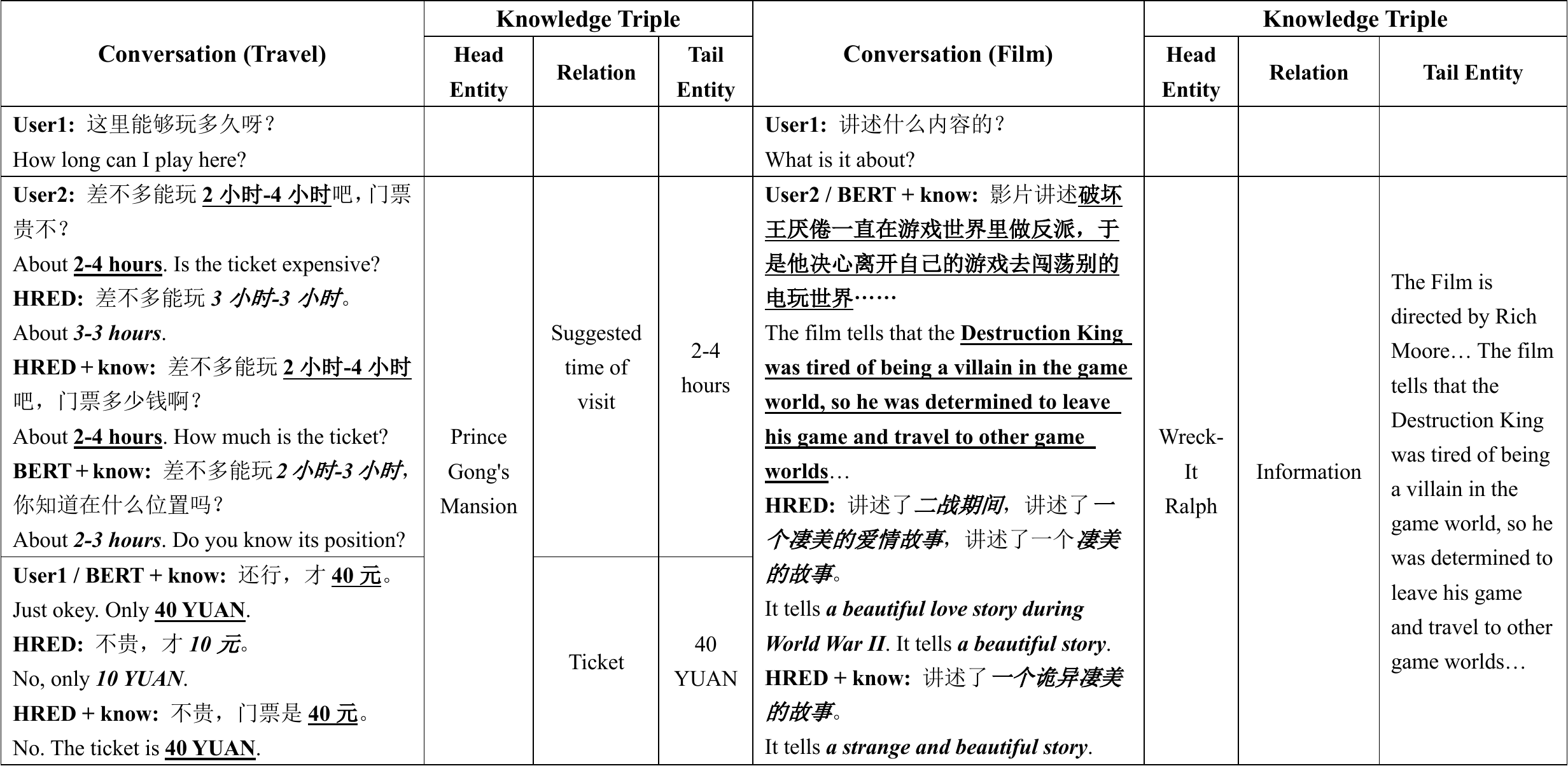}
\caption{Two cases of the \textbf{travel} and \textbf{film} domains. 
The \underline{\textbf{underlined}} text is the knowledge used by the golden truth or the knowledge correctly utilized by the models. The \textit{\textbf{italic}}
text are contradictory to the background knowledge.
}
\label{fig:case}
\end{figure*}

\subsubsection{Metrics}
Human annotators were asked to score a generated response in terms of the fluency and coherence metrics. 

\noindent \textbf{Fluency} (rating scale is 0,1,2) is defined as whether the response is fluent and natural: 
\begin{itemize}
\item score 0 (bad): it is not fluent and difficult to understand due to grammatical errors.
\item score 1 (fair): it contains some grammatical errors but is still understandable.
\item score 2 (good): it is fluent and plausibly produced by a human.
\end{itemize}

\noindent \textbf{Coherence} (rating scale is 0,1,2) is defined as whether a response is relevant and coherent to the context and the knowledge information:
\begin{itemize}
\item score 0 (bad): it is irrelevant to the context.
\item score 1 (fair): it is relevant to the context but not coherent to the knowledge information.
\item score 2 (good): it is both relevant to the context and coherent to the background knowledge.
\end{itemize}

\subsubsection{Annotation Statistics}
We randomly sampled about 500 contexts from the test sets of the three domains and generated responses by each model. These 1,500 context-response pairs in total and related knowledge graphs were presented to three human annotators.

We calculated the Fleiss' kappa \cite{fleiss1971measuring} to measure inter-rater consistency. Fleiss' kappa for Fluency and Coherence is from 0.37 to 0.74, respectively. The overall 3/3\footnote{3/3 means all the three annotators assign the same label to an annotation item.}
agreement for Fluency and Coherence is from 68.14\% to 81.33\% in the three domains.

\subsubsection{Results}

The results are shown in Table \ref{tab:manual}. As can be seen, knowledge-aware BERT outperforms other models significantly in both metrics in all the three domains, which agrees with the results of automatic evaluation. The Fluency is 2.00 because the retrieved responses are all human-written sentences. The Fluency scores of both generation-based models are close to 2.00 (in the music domain, the Fluency of HRED is 1.90), showing that the generated responses are fluent and grammatical. 
The Coherence scores of both HRED and knowledge-aware HRED are higher than 1.00 but still have a huge gap to 2.00, indicating that the generated responses are relevant to the context but not coherent to knowledge information in most cases. After incorporating the knowledge information into HRED, the Coherence score is improved significantly in all the three domains, as the knowledge information is more expressed in the generated responses.

\subsection{Case Study}

Some sample conversations in the travel and film domains are shown in Figure \ref{fig:case}. As we can see, HRED tends to generate responses which are relevant to the context, while incoherent with the knowledge base. After introducing knowledge information, HRED is able to generate knowledge-grounded responses, for instance, the replies of HRED with the knowledge in the travel domain. However, generating knowledge-coherent responses with reference to unstructured text knowledge is still difficult for knowledge-aware HRED (see the conversation in the film domain), as modeling the knowledge of unstructured texts requires more powerful models. For knowledge-aware BERT, the retrieved responses are coherent with the context and the knowledge information in most cases. However, it may focus on the semantic information of conversations but ignore the knowledge information, as shown in the conversation in the travel domain, which may be addressed by knowledge-enhanced pre-trained models, like ERNIE \cite{sun2019ernie}.

\section{Conclusion and Future Work} 
\label{sec:conclusion}

In this paper, we propose a Chinese multi-domain corpus for knowledge-driven conversation generation, KdConv. It contains 86K utterances and 4.5K dialogues, with an average number of 19.0 turns. Each dialogue contains various topics and sentence-level annotations that map each utterance with the related knowledge triples. The dataset provides a benchmark to evaluate the ability to model knowledge-driven conversations. In addition, KdConv covers three domains, including film, music, and travel, that can be used to explore domain adaptation or transfer learning for further research. We provide generation- and retrieval-based benchmark models to facilitate further research. Extensive experiments demonstrate that these models can be enhanced by introducing knowledge, whereas there is still much room in knowledge-grounded conversation modeling for future work.

\section*{Acknowledgments}
This work was jointly supported by the NSFC projects (Key project with No. 61936010 and regular project with No. 61876096), and the National Key R\&D Program of China (Grant No. 2018YFC0830200). We thank THUNUS NExT Joint-Lab for the support.

\bibliography{anthology,acl2020}
\bibliographystyle{acl_natbib}

\end{document}